\documentclass{article}
\usepackage[english]{babel}

\usepackage[letterpaper,top=2cm,bottom=2cm,left=3cm,right=3cm,marginparwidth=1.75cm]{geometry}

\usepackage{amsmath}
\usepackage{paralist}
\usepackage{multirow}
\usepackage{arydshln}
\usepackage{color,soul}
\usepackage{setspace}
\usepackage{gensymb}
\usepackage[affil-it]{authblk}
\usepackage{graphics} 
\usepackage{epsfig} 
\usepackage{mathptmx} 
\usepackage{times} 
\usepackage{amsmath,amssymb,amsfonts}
\usepackage{algorithmic}
\usepackage{bm}
\usepackage{cite}
\usepackage{gensymb}
\usepackage{adjustbox}
\usepackage{graphicx}
\usepackage{soul}
\usepackage[subnum]{cases}
\usepackage{textcomp}
\usepackage{wasysym}
\usepackage[english]{babel}
\usepackage{multirow}
\usepackage[colorlinks=true, allcolors=blue]{hyperref}

\title{Real-time, accurate, and open source upper-limb musculoskeletal analysis using a single RGBD camera}

\author[1,*]{Amedeo Ceglia}
\author[2]{Kael Facon}
\author[1,3]{Micka\"el Begon}
\author[4]{Lama Seoud}

\affil[1]{Institute of Biomedical Engineering, University of Montreal, Montreal, QC Canada}
\affil[2]{Ecole d'ingénieurs informatique Paris, Le Kremlin-Bicêtre, France}
\affil[3]{School of Kinesiology and Human Kinetics,
University of Montreal, Montreal, QC Canada}
\affil[4]{Department of Computer Engineering and Software Engineering, Polytechnique Montreal, Montreal, QC, Canada}
\affil[*]{Corresponding author: Amedeo Ceglia, amedeo.ceglia@umontreal.ca}

\providecommand{\keywords}[1]
{
  \small	
  \textbf{\textit{Keywords---}} #1
}
\date{}

\begin{document}
\maketitle
\begin{abstract}
Biomechanical biofeedback may enhance rehabilitation and provide clinicians with more objective task evaluation. These feedbacks often rely on expensive motion capture systems, which restricts their widespread use, leading to the development of computer vision-based methods. These methods are subject to large joint angle errors, considering the upper limb, and exclude the scapula and clavicle motion in the analysis. Our open-source approach offers a user-friendly solution for high-fidelity upper-limb kinematics using a single low-cost RGBD camera and includes semi-automatic skin marker labeling. Real-time biomechanical analysis, ranging from kinematics to muscle force estimation, was conducted on eight participants performing a hand-cycling motion to demonstrate the applicability of our approach on the upper limb. Markers were recorded by the RGBD camera and an optoelectronic camera system, considered as a reference. Muscle activity and external load were recorded using eight EMG and instrumented hand pedals, respectively. Bland-Altman analysis revealed significant agreements in the 3D markers' positions between the two motion capture methods, with errors averaging 3.3$\pm$3.9\,mm. For the biomechanical analysis, the level of agreement was sensitive to whether the same marker set was used. For example, joint angle differences averaging 2.3$\pm$2.8\degree~when using the same marker set, compared to 4.5 pm 2.9\degree~otherwise. Biofeedback from the RGBD camera was provided at 63\,Hz. Our study introduces a novel method for using an RGBD camera as a low-cost motion capture solution, emphasizing its potential for accurate kinematic reconstruction and comprehensive upper-limb biomechanical studies.
\end{abstract}

\section{Introduction}
Real-time biomechanical feedback is highly relevant to improve or guide rehabilitation \cite{giggins2013biofeedback}. Feedbacks may include joint kinematics \cite{bigoni2016does}, joint kinetics \cite{femery2004real}, or muscle activity \cite{bolek2003preliminary}.
Joint kinematics is often required for subsequent biomechanical analyses and relies on fast and straightforward algorithms (e.g., inverse kinematics) based on motion capture data.\\
Motion capture is commonly based on optoelectronic camera systems to track reflective or active markers placed on bony landmarks. This technology is accurate, with marker position errors of less than a millimeter \cite{topley2020comparison}, making it robust for clinical application and a reference for assessing emerging motion capture technologies. 
However, using optoelectronic systems in clinical or sports environments is nevertheless challenging due to their high cost and setup complexity. Consequently, researchers have developed affordable alternative systems to address these limitations. Three technologies stand out as the most used alternatives.\\
The first technology is based on inertial measurement units (IMUs) placed on predefined positions of body segments \cite{niswander2020optimization}. IMUs have the advantage of being compatible with clothing in nearly any situation and suitable for use in confined environments. However, the gyroscope may experience drift during long tasks, and the electromagnetic fields of the magnetometer can be affected, leading to errors in joint angle estimates around 10\degree~for the lower limb \cite{shuai2022reliability} and up to 30\degree~for the upper limb \cite{henschke2022assessing} when compared to marker-based systems. Another limitation is the oversimplification of the upper-limb model with a single humerothoracic joint instead of modeling the sternoclavicular, acromioclavicular, and glenohumeral joints. The role of the scapula, which holds significance in upper limb rehabilitation \cite{voight2000role} is consequently omitted.\\
The second technology employs a set of synchronized and calibrated RGB cameras combined with deep learning algorithms to recognize 3D human poses without markers. The algorithm detects a series of key points representing joint centers by inferring 3D pose from 2D pose provided by each calibrated camera \cite{lahkar2022accuracy}. These key point 3D positions are then used to compute the joint angles. Such markerless systems reduce experimental time and are less intrusive, as they can be used while wearing clothes. However, similar to marker-based systems, implementing a camera-based system can be costly, require large space, and achieving a precise setup can be challenging.
All cameras must be synchronized and calibrated together, as the final 3D pose accuracy relies on multi-view geometry. Additionally, pose estimation is resource-demanding and requires high GPU hardware, making it unsuitable for real-time biofeedback or embedded setups.
This approach is promising but relies on annotated images on which users have limited access to the reliability of the annotations. Depending on the specific use case, the accuracy may be insufficient, particularly for internal/external arm rotation with error reaching up to 23\degree~\cite{lahkar2022accuracy}. Moreover, like IMU-based technologies, models only include a simplistic humerothoracic joint (excluding the clavicle and scapula).\\
The last technology is based on a single camera (with or without depth measurements) combined with machine learning, to locate joint centers and compute joint angles using pre-trained data, like in \cite{boldo2024reliability}. No external camera calibration or synchronization is needed as only one camera is used, making it a cost-effective and user-friendly approach. This technology shares similar drawbacks as the previous approach, including inaccuracies in annotated images and errors in localizing the joint center, potentially resulting in an error of up to 25\degree~for upper limb movements \cite{sarsfield2019clinical}. Moreover, when the depth measurement is not available, the input information (2D image) is limited, requiring the algorithm to make numerous assumptions to infer the 3D pose estimation. As with the two last technologies, the model is also simplified for the shoulder joint. 
Given that, some methods, such as those using optical-depth cameras (RGBD), have shown promising results in providing low-cost and real-time marker-based motion capture systems. It is noteworthy that, to the best of our knowledge, these methods have not been applied to a biomechanical analysis of the upper limb. In \cite{ye2016depth}, the use of infrared images and reflective markers has been observed to reduce the accuracy of depth information around the markers, while in \cite{timmi2018accuracy}, the algorithm’s robustness requires a spherical marker with a large radius, limiting its applicability to upper limb motion capture.
In this study, we introduced an open-source and user-friendly method using a single low-cost RGBD camera and 13 flat white skin markers to record the upper limb kinematics. Our primary goal was to develop a simple, rapid, and cost-effective system for capturing kinematic data in clinical environments. We implemented a semi-automatic marker labeling algorithm and established a biomechanical framework based on a musculoskeletal model.
We conducted a comparison of our method against a marker-based approach, considered as the reference. Our evaluation encompassed the 3D positions of the markers, biomechanical outcomes, as well as time performances. The objective was to offer a comprehensive solution for upper limb motion capture, providing a balance between accuracy and ecological usability.

\section{Method}
\subsection{Data collection}
\begin{figure}
\centering
    \includegraphics[width=0.8\textwidth]{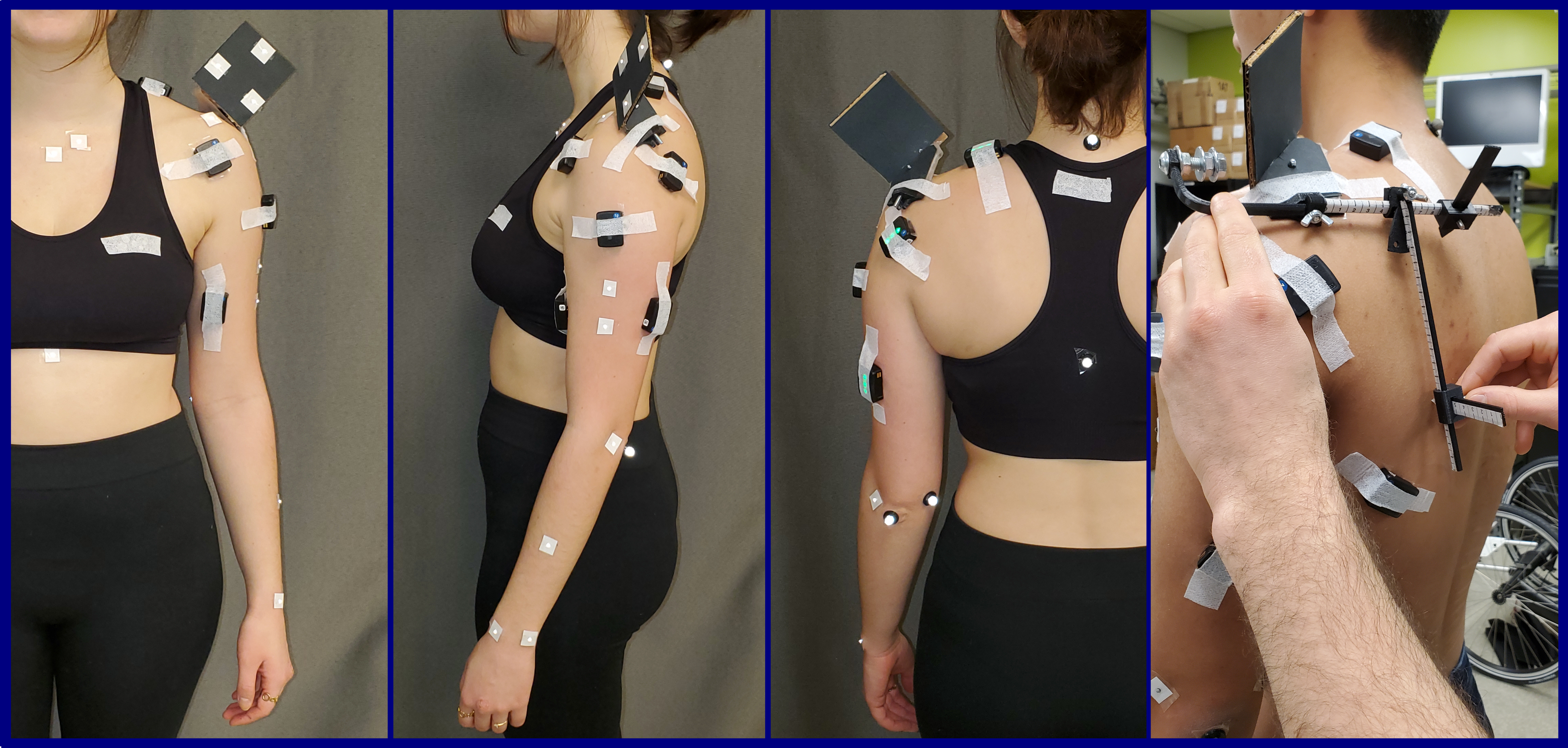}
    \caption{Participant outfitted with white markers, reflective markers (center of the square), and EMG sensors on the left arm. Manual calibration of the acromion cluster \cite{ceglia2024sofamea} is shown on the right. The scales on the device facilitate the extraction of anatomical bony landmarks from the 3D cluster's marker positions via an included Python code.}
    \label{fig:participant}
\end{figure}

\begin{figure}
\centering
    \includegraphics[width=0.8\textwidth]{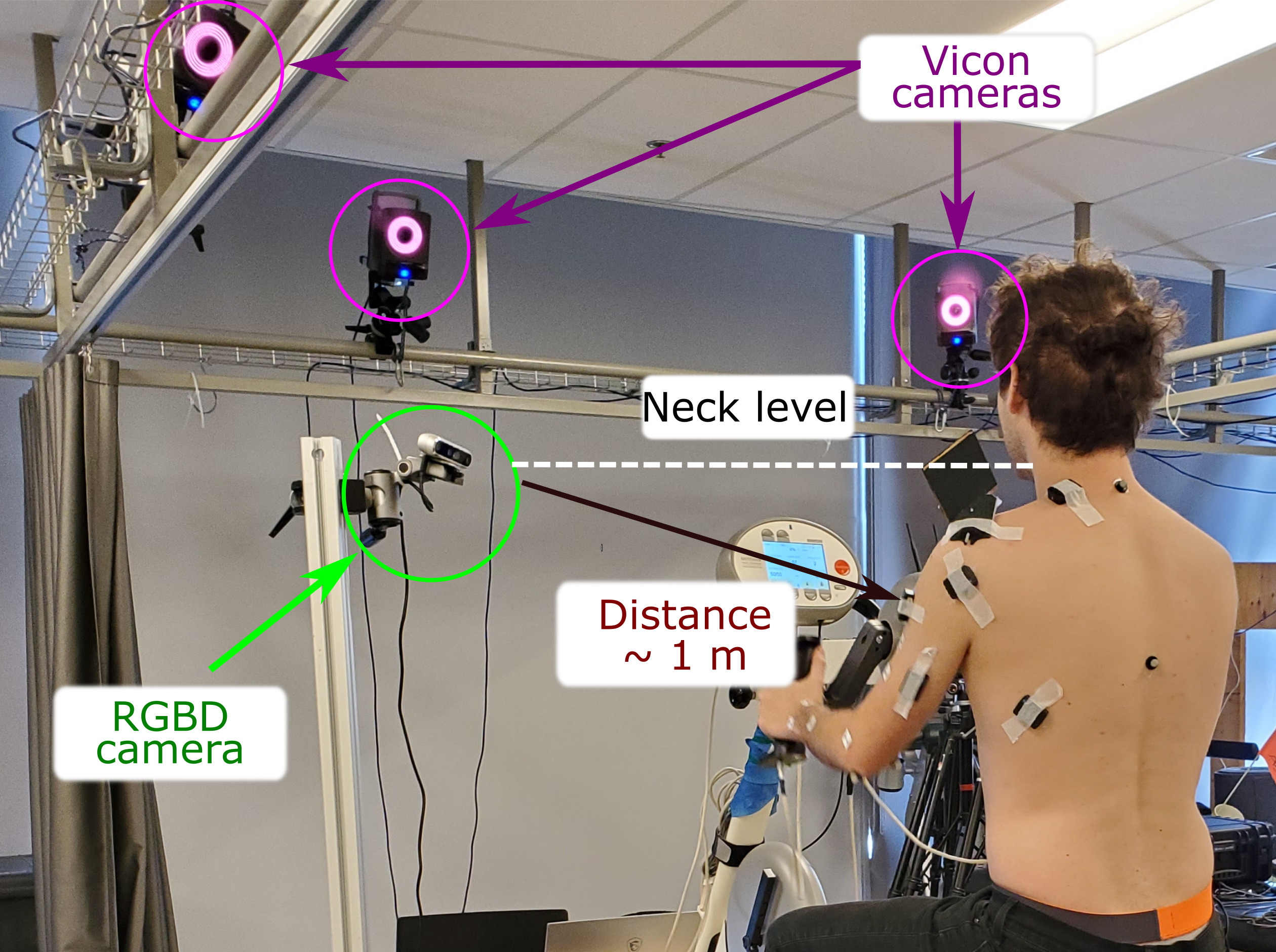}
    \caption{Experimental setup showing the RGBD camera placement and some Vicon cameras while a participant performs hand cycling.}
    \label{fig:setup}
\end{figure}

Eight healthy participants (3 females; age: 25.9\,$\pm$\,12.2\,years old; height: 175.8\,$\pm$\,8.5\,cm; mass: 70.2\,$\pm$\,8.4\,kg) were involved after signing an informed consent. This experiment was approved by the ethics committee of the Université de Montréal (\#2023-4743). 
A single stereo-vision RGBD camera (RealSense D455, Intel, USA) was used to record both color (RGB) and depth image frames at 60\,Hz with a 848x480 pixel resolution. It was positioned approximately one meter from the thorax, roughly aligned with the participant's neck. Thirteen squared white markers ($\sim$2$\times$2\,cm) were placed on the trunk and left upper limb at predefined locations to detect key bony landmarks and reduce soft tissue artifacts \cite{blache2017main} (Fig.~\ref{fig:participant}). A 14-camera motion capture system (T40s, Vicon, Oxford, UK) was used to record (120\,Hz) 13 hemispherical (\diameter 1\,mm) retro-reflective markers placed in the center of the white markers, and 5 additional \diameter 10\,mm reflective markers placed on the back of the trunk and the medial side of the arm (Fig.~\ref{fig:participant}). To fairly compare both methods, we placed more reflective markers to further evaluate the impact of the markers' redundancy, as the optoelectronic system can easily acquire numerous markers. The term "minimal-Vicon-based" refers to using the same marker set as the RGBD-based method, while "redundant-Vicon-based" denotes the use of additional markers. As the markers on the back were not available with the reduced marker set, measurements were manually taken between the 7th cervical vertebra (C7) and the manubrium using an anthropometric caliper for a subsequent model scaling.
The participants were equipped with a validated custom-made acromion marker cluster \cite{ceglia2024sofamea}, enabling the estimation of the scapula's bony landmarks (acromial angle, trignium spinae, and inferior angle) from a front view without the need for optical calibration.
Muscle activities were recorded at 2160\,Hz using eight surface EMG sensors (Trigno EMG Wireless System, Delsys, USA) put on the Pectoralis (major), Biceps brachii (long head), Triceps brachii (long heads), Upper Trapezius, Deltoid (anterior, medial, posterior heads), and Latissimus Dorsi, after shaving and cleaning the skin.
The hand cycle was equipped with instrumented hand pedals (Sensix, France) that recorded the 3D forces and moments at 100\,Hz. The experimental setup is shown in Fig.\,\ref{fig:setup}.
EMG and reflective markers data were synchronized using Nexus 2.11 (Vicon, Oxford, UK). A trigger signal was recorded on Vicon and used in concurrency with Biosiglive \cite{ceglia2023biosiglive} to start the recording of the RGBD and the hand pedals at the same time.
Participants were asked to complete two 5-second hand-cycling trials at maximal effort, maintaining a cadence of 60 RPM, with a 2-minute rest interval in between. The resulting signals were used to normalize EMG for subsequent muscle force estimation. 
Finally, the participants performed one anatomical pose (for musculoskeletal model scaling) and 2\,min-hand pedal trials at 15, 20, 30, and 40\,W all performed at 60\,RPM.

\begin{figure}
\centering
    \centerline{\includegraphics[width=\columnwidth]{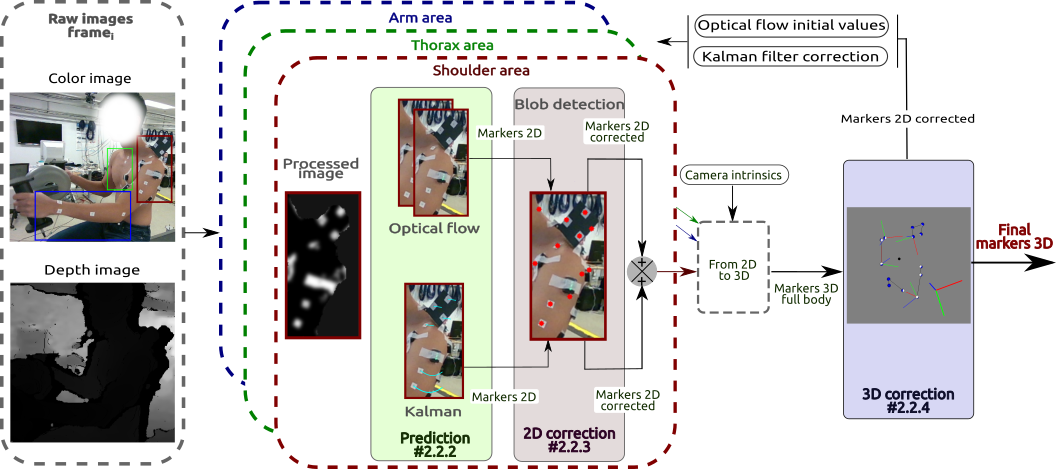}}
    \caption{Software architecture of the markers tracking from the RGBD camera.}
    \label{fig:pipeline_label}
\end{figure}

\subsection{Marker trajectories extraction from RGBD camera}
\label{autolabeling_markers}
The \textit{RGBDMocap} Python package \cite{rgbdmocap} was developed to detect and label the markers in RGB images ($x_c$ and $y_c$ marker positions) and then retrieved their corresponding depth ($z_c$) positions. The color and depth images were aligned using the pre-implemented RealSense\textregistered~algorithm to establish corresponding depth values for each color pixel. The markers' labeling was implemented via a four-step workflow to ensure detection in each frame (Fig.\,\ref{fig:pipeline_label}), which is described thereafter:
\begin{inparaenum}[1)]
\item initialization (once on the first frame);
\item markers 2D prediction;
\item markers 2D correction;
\item markers 3D correction.
\end{inparaenum}

\subsubsection{Initialization}
\label{initialization}
\begin{figure}
\centering
    \includegraphics[width=1\textwidth]{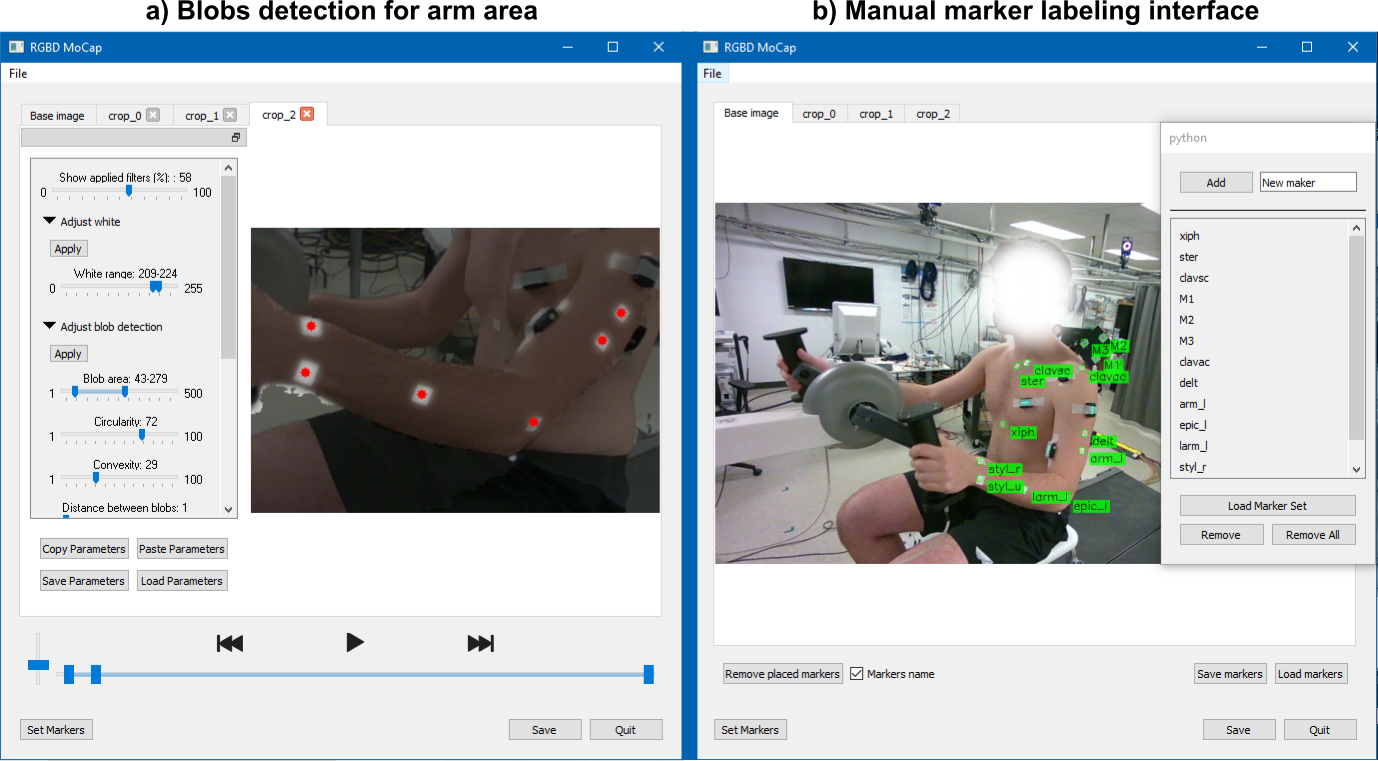}
    \caption{Graphical user interface for a) blob detection for the arm area and b) manual markers labeling.}
    \label{fig:gui}
\end{figure}

To provide robust, rapid, and resource-efficient detection, the images underwent segmentation into three distinct areas: thorax, shoulder, and arm. 
We treated the three areas in parallel using the Python multiprocessing package \cite{hunt2019multiprocessing}.
All ensuing initialization procedures were individually applied to each area. 
A graphical user interface (GUI) was developed to facilitate the process (Fig.\,\ref{fig:gui}).
Filters were applied to the images to improve the white markers detection on the body, reflecting the initial tuning of a set of parameters for each optoelectronic camera.
Specifically, the background was first removed using the aligned depth image. A blur filter was then applied to enhance the image by smoothing its edges.
Contrast-limited adaptive histogram equalization was then applied using the OpenCV function \cite{howse2013opencv} (the threshold for contrast limiting and the size of tiles in the row and column were chosen empirically for each area). Finally, a binary threshold was applied to increase white markers' visibility. 
The blob (i.e., region of an image in which some properties are approximately constant) detector from the OpenCV library was employed to detect the white square markers automatically. This detector identifies groups of connected pixels using criteria such as white range color, area, circularity, and convexity. 
These criteria were manually adjusted in the GUI (Fig.\,\ref{fig:gui}-a)) for each area and participant to provide the best blob detection. Once the camera was positioned and the parameters determined for the first participant, they yielded reasonably good results for the subsequent participants. Nevertheless, fine-tuning for each participant is recommended.
Afterward, the user labeled each white marker once, forming the initial value for the following prediction step (Fig.\,\ref{fig:gui}-b)).
Once the initialization was done, the three-step method was used for each frame.

\subsubsection{Marker 2D prediction}
\label{markers_2d_prediction}
As the movement induced noise and lighting variations in the images, the blob detector may detect false positive blobs or miss some white markers in the image. Therefore, it was not sufficient to detect the blobs and track the trajectory by finding the closest neighbor. A predictive step using the markers' previous positions was conducted to estimate the subsequent positions of the markers.
Two predictive methods were used independently to improve the robustness of our algorithm: 
\begin{itemize}
     \item An \textbf{optical flow detector} tracks each previously labeled marker by analyzing the differences between the previous and current images. A sparse iterative version of the Lucas-Kanade method in pyramids \cite{bouguet2001pyramidal} implemented in OpenCV was used. This algorithm provides an estimate of the position of a specific pixel by tracking the change of intensity of a pixel over time. Initial labeled marker positions were used as the initial positions.
     \item The \textbf{Kalman filter} predicts the 2D positions of the markers, leveraging their previous positions and velocities. This method exclusively relies on the positions of detected blobs. By employing a covariance matrix, the filter predicts the current position. The implementation of the Kalman filter was achieved through OpenCV, employing marker position and velocity as states. The initial positions of the labeled markers serve as the filter initial states. 
 \end{itemize} 

\subsubsection{Markers 2D correction}
\label{markers_2d_correction}
The estimated marker positions were compared to the detected blob positions within the image. If the distance between the predicted position and the detected blob was less than 10\,pixels, the marker was classified as \textit{visible}, and its position was replaced by the blob's position. Otherwise, the estimated position remained unchanged, and the marker was classified as \textit{not visible}. On average, the marker size was about 10$\times$10\,pixels at 1\,m from the camera. 
The marker positions generated by both tracking methods were combined by averaging the positions of visible markers. In the case where neither method detected a visible marker, its last known position was retained for the next step.
While this approach performs well when markers are visible, it encounters issues when blobs are not consistently detected, leading to uncertainty. Consequently, an additional correction step was implemented in the 3D space using markers altogether to mitigate these issues. 

\subsubsection{Markers 3D correction}
\label{markers_3D_correction}
The previous procedures (Sec.\,\ref{markers_2d_prediction}, \ref{markers_2d_correction}) enabled marker tracking exclusively through 2D color images. However, this section leverages the complete capabilities of the RGBD camera by operating within the 3D camera space. To achieve this, a 3D articulated model was built from the initial marker 3D positions, consisting of four segments (thorax, shoulder, arm, and lower arm) connected by ball-and-socket joints, except for the thorax which was free in space.
The marker 2D positions from the previous steps (Sec. \ref{markers_2d_prediction}-\ref{markers_2d_correction}), coupled with the depth image, were used to retrieve the 3D marker positions in the camera coordinate system using the default RGBD-camera intrinsic parameters.
An inverse kinematics algorithm based on an Extended Kalman Filter (EKF) \cite{fohanno2010estimation} was then applied to the 3D marker positions to estimate the model joint angles, using the Biorbd library \cite{michaud2021biorbd}. The EKF is a common sensor fusion method that aims to estimate the state (here joint angles and velocities) of a non-linear system given noisy measurements (here the markers) by minimizing the trace of the error covariance matrix. 
These joint angles provided an estimate of the position of the model and segment in the camera 3D space. Forward kinematics was then used to retrieve marker positions from joint angles. These 3D positions were re-projected into image coordinates using the RealSense\textregistered~re-projection function along with intrinsic parameters.
Finally, the same 2D correction step described in Sec.\,\ref{markers_2d_correction} was performed to find the corrected 2D markers from forward kinematics based on blobs detection. 
Subsequently, these final 2D marker positions were used for the Kalman correction and the optical flow input positions to provide a corrected initial value for the markers' prediction on the next frame. 

\subsection{RGBD - Vicon registration}
Section \ref{autolabeling_markers} allowed us to retrieve marker positions in the camera coordinate system ($x_c$, $y_c$, $z_c$). 
To enable a meaningful comparison between 3D positions obtained via the RGBD and those provided by Vicon-based mocap, it was necessary to express the 3D positions from both systems in the same coordinate system.
The marker 3D positions assessment was conducted within the camera coordinate system ($x_c$, $y_c$, $z_c$) to assess the error along each camera axis since color image ($x_c$, $y_c$) and depth image ($z_c$) come from different sensors.
For the biomechanical analysis, positions were defined within the Vicon coordinate system ($x_v$, $y_v$, $z_v$), defined by a vertically oriented z-axis.
The coordinate system transformations were made through an optimal transformation matrix that aimed to minimize the markers' error on the initial pair of synchronized frames.
Subsequently, the RGBD-to-Vicon roto-translation matrix was applied to transform the 3D position of all RGBD markers across all frames. Similarly, the inverse of this matrix was used to transform each Vicon marker into the RGBD coordinate system. Additionally, RGBD recording may not consistently maintain a 60\,Hz rate, as indicated by the discontinuous camera frame number provided by the camera. These gaps in frame numbers were used to interpolate data at 60\,Hz to fill the missing frames. Then, linear interpolation was used to match the frame rate of the Vicon system (120\,Hz). This approach facilitated the comparison of time series data between the RGBD and Vicon systems.
The ensuing section (Sec.\,\ref{biomec_analys}) employs the modified RGBD marker positions expressed in the Vicon coordinate system.

\subsection{Real-time biomechanical framework}
\label{biomec_analys}
A real-time biomechanical framework was developed and applied to the collected data using the two motion capture systems considering the two marker sets (i.e. RGBD, minimal-Vicon, and redundant-Vicon).
The framework included inverse kinematics to compute joint angles from marker positions, inverse dynamics to calculate joint torques, and static optimization to estimate muscle forces.

\subsubsection{Musculoskeletal model}
The musculoskeletal model, based on Wu et al.'s upper-limb model \cite{wu2016subject}, encompasses five segments (thorax, clavicle, scapula, arm, and lower arm) with 10~degrees of freedom (two sternoclavicular, three acromioclavicular, and three glenohumeral rotations, along with elbow flexion and pronation/supination).
A static trial was used to scale the model for each motion capture method, resulting in three separate scaled models. The thorax width (anteroposterior) was scaled using measurements taken between the manubrium and the C7. 
The model is actuated by 31 Hill-type muscles, including contractile and passive elements in parallel.
To speed up computations, the tendons were considered rigid \cite{millard2013flexing}. Moreover, as the movements of interest were slow (60\,RPM), muscles were directly controlled by their activation (neglecting neural excitation dynamics with its electromechanical delay \cite{neptune2001muscle}).
For a muscle $m$, the force ($f^{m}$) computed from activation ($a$) and muscle kinematics (normalized muscle fiber length $\tilde{l}_m$ and velocity $\tilde{v}_m$) was expressed as:
\vspace{-0.2cm}
\begin{equation}
 f^{m} = f^{m}_{o}(af^{act}(\tilde{l}_{m})f^{v}(\tilde{v}_{m})+f^{pas}(\tilde{l}_{m}))\cos \alpha, \label{mus_force_eq}
\vspace{-0.26cm}
\end{equation}
where $f^{m}_{o}$ is the muscle maximal isometric force, 
$f^{act}(\tilde{l}_{m})$, $f^{v}(\tilde{v}_{m})$, and $f^{pas}(\tilde{l}_{m})$ are the active force-length, force-velocity, and the passive force-length relationships, respectively; $\alpha$ is the muscle pennation angle \cite{de2016evaluation}.

\subsubsection{Inverse kinematics}
\label{inverse_kin}
First, the trajectories of the markers were smoothed by applying a moving average window spanning 14 frames (7 frames delay). Then, the inverse kinematics was performed using an EKF \cite{fohanno2010estimation} implemented in the Biorbd software, with the joint angles $q$, velocity $\dot{q}$ and, acceleration $\Ddot{q}$ as the states. 
The reader can refer to \cite{joukov2020real} for further information on inverse kinematics using EKF. 

\subsubsection{Inverse dynamics}
\label{inverse_dyn}
The joint torques ($\tau$) were calculated from joint kinematics ($q$, $\dot{q}$, $\Ddot{q}$ outputting from the EKF), external forces ($F_{ext}$) measured by pedal force sensors and segment inertial parameters (encompassed in the model):
\begin{equation}
\label{eq_motion}
    \tau = M(q)\Ddot{q} + N(q, \Dot{q}) + G(q) + C^T\!(q)F_{ext}
\end{equation}
\noindent where $M(q)$, $N(q, \Dot{q})$, $G(q)$, and $C(q)$ are the mass matrix, the vector of Coriolis and centrifugal forces, the vector of gravity effects, and the Jacobian matrix of the hand force point of application, respectively. 

\subsubsection{EMG-informed static optimization}
\label{static_optim}
Static optimization for estimating muscle forces corresponded to a least square problem aiming to find the optimal muscle activations ($\boldsymbol{a}$) according to the joint torque, provided through inverse dynamics, and the experimental EMG. 
Offline, before the hand cycling trials, EMG signals of maximal hand-cycling trials were band-pass filtered (10-425\,Hz), rectified, and low-pass filtered (5\,Hz) using a fourth-order Butterworth filter. Then, activation for each recorded muscle during the maximal effort trials was obtained by computing the mean of the highest values for 1\,s. 
The static optimization was defined as: 
\vspace{-0.2cm}
\begin{subequations}
\begin{equation}
\begin{split}
\min_{a,~\tau^{res}}~\sum \limits_{k=1}^{n_{emg}} \omega_{emg} \|emg_k - a^{inf}_k\|^{2} 
+ \sum \limits_{i=1}^{n_{muscles} - n_{emg}} \omega_a\|a^{ninf}_i\|^{2} \\
+ \sum \limits_{j=1}^{n_{\text{joints}}}\omega_{\tau} \|\tau_j - (\tau^{m}_j + \tau^{res}_j)\|^{2} + \omega_{res}\|\tau^{res}_j\|^{2} 
\end{split}
\label{eq:cost}
\vspace{-1.5cm}
\end{equation}
\begin{equation}
\forall t, ~\boldsymbol{a} \in \mathcal{A},~\boldsymbol{\tau^{res}} \in \mathcal{T} \label{eq:bounds}
\end{equation}
\label{eq:optim}
\vspace{-0.5cm}
\end{subequations}

\noindent where $a^{inf}$ and $a^{ninf}$ are the informed and non-informed muscle activations, respectively. $\tau$ corresponds to the joint torques obtained by inverse dynamics Eq.\,\ref{eq_motion}, $\tau^{res}$ is a residual joint torque modeling the inner stiffness of the joint, and $\tau^{m}$ are the muscle torques computed through Eq.\,\ref{mjt}. $emg$ stands for the eight recorded EMG signals. $\omega_a$, $\omega_{emg}$, $\omega_{\tau}$, and $\omega_{res}$ represent the weights assigned to minimize muscle activations, track experimental EMG, track inverse dynamics torques, and minimize residual torque, respectively. Eq.~\ref{eq:bounds} ensures the physiological bounds of the muscle activation between 0 and 1 and bound the residual torque between -5\,N.m and 5\,N.m
The joint torque from the muscles $m$ crossing the joint $j$ is: 
\vspace{-0.2cm}
\begin{equation}
    \tau^m_j = \sum \limits_{i=1}^{n_{\text{muscles}}} \Gamma_i \cdot f^m_i 
    \label{mjt}
\end{equation}
\noindent where $\Gamma_i$ is the moment arm of the muscle $i$ calculated as the Jacobian matrix of the muscle length.
Despite the existing boundaries, no additional constraints were incorporated into the optimization process to ensure successful solving. However, the weight on joint torque tracking must be set sufficiently high to guarantee faithful tracking.
This optimization was performed using Acados \cite{verschueren2022acados} leveraging the efficient implementation of the linear solver Blasfeo and the QP solver OSQP \cite{stellato2020osqp} allowing a fast solving.


\subsection{Data analysis}
The assessment of the \textit{RGBDmocap} system focused on two key aspects: 1) the accuracy of marker positions and error propagation through the biomechanical analysis outcomes compared to the Vicon system, and 2) the overall biofeedback latency.
To begin with, the comparative analysis between the two systems was conducted through three distinct methods: \begin{inparaenum}[1)]
\item Calculation of the absolute errors between the positions of RGBD-based markers and those derived from the Vicon-based method;
\item Examination of error propagation within the biomechanical study outcomes;
\item Investigation into the impact of the marker redundancy in the RGBD-based method on biomechanical results. 
\end{inparaenum}
The agreement between 3D marker positions obtained from the two different methods, namely Vicon and RGBD, was evaluated within the RGBD camera coordinate system through Bland–Altman analysis \cite{bland2007agreement}. Specifically, the \textit{bias} and the \textit{limits of agreement} (LoA) were used to assess the agreement. Within the analysis, Bland-Altman \textit{differences} were computed along the mean of the $x_c$ and $y_c$ axes corresponding to the error in the color image plan, and along the $z_c$ axis corresponding to the error due to the depth values. Bland-Altman \textit{mean} values were calculated along the $z_c$ axis, aiming to explore the impact of the markers-to-camera distance.
The same analysis was applied to the biomechanical study outcomes, encompassing joint angles, joint torques, and muscle forces. Here, the agreement was assessed by comparing both RGBD-based and minimal-Vicon-based methods against the redundancy-Vicon-based method (our reference), and by comparing the RGBD-based method to the minimal-Vicon-based method. This approach resulted in three distinct analyses aimed at evaluating agreement between methods and assessing the dependency of results on the number of acquired markers. The following notations were used to ease the reading: RGBD vs redundant-Vicon (RGBDvRED), minimal-Vicon vs redundant-Vicon (MINvRED), and RGBD vs minimal-Vicon (RGBDvMIN)\\
To be applicable in a clinical context, feedback must be provided in real-time and with low latency. To assess the real-time capability of the RGBD-based method, the delay between motion and outcomes was evaluated. This delay contains the time required to retrieve the labeled position and to compute the biomechanical outcomes for each frame. The delay induced by the marker filtering (Sec.\,\ref{inverse_kin}) was also considered. The tracking algorithm and biomechanical analysis were performed on a laptop (Intel\textregistered CoreTM i7-6700HQ CPU @ 2.60 GHz), operating with Ubuntu 20.04 LTS. 

\section{Results}
\subsection{3D Markers positions}
\begin{figure}
\centering
    \centerline{\includegraphics[width=\columnwidth]{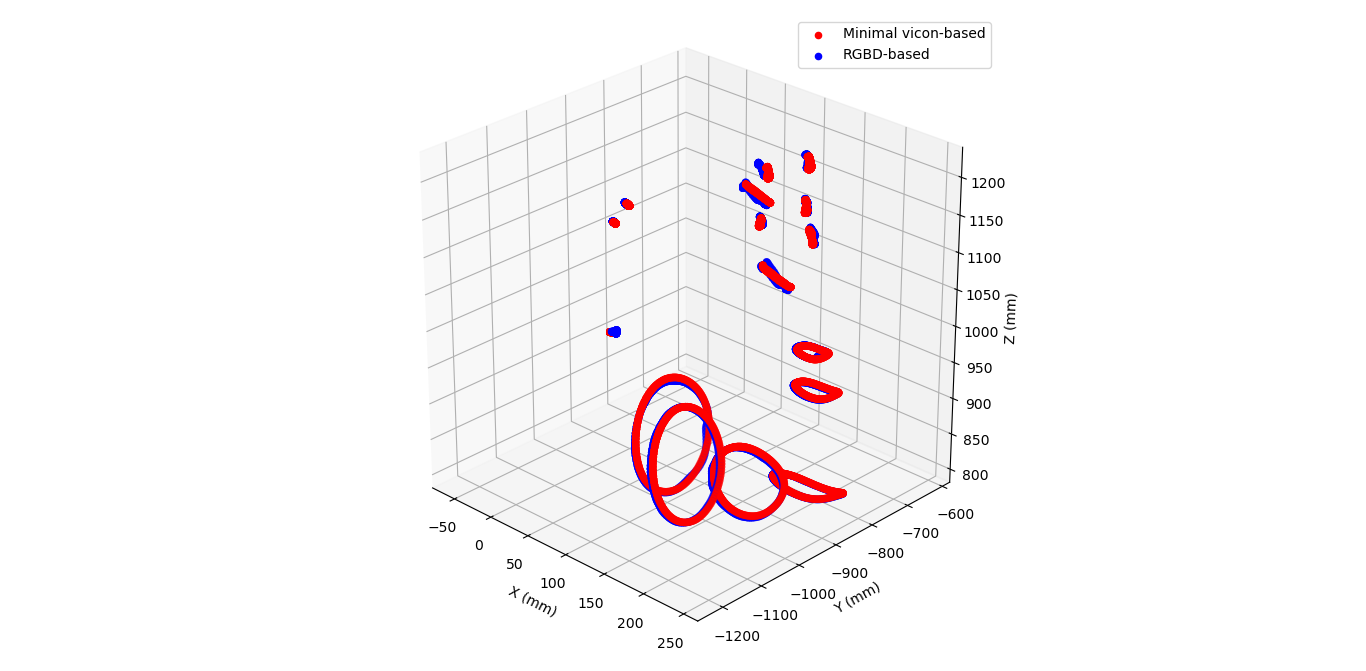}}
    \caption{Mean of the 120 cycles of 3D marker trajectories for one participant for minimal-Vicon (red) and RGBD-based (blue) mocap methods.}
    \label{fig:markers_3d}
\end{figure}

\begin{figure}
\centering
\centerline{\includegraphics[width=\columnwidth]{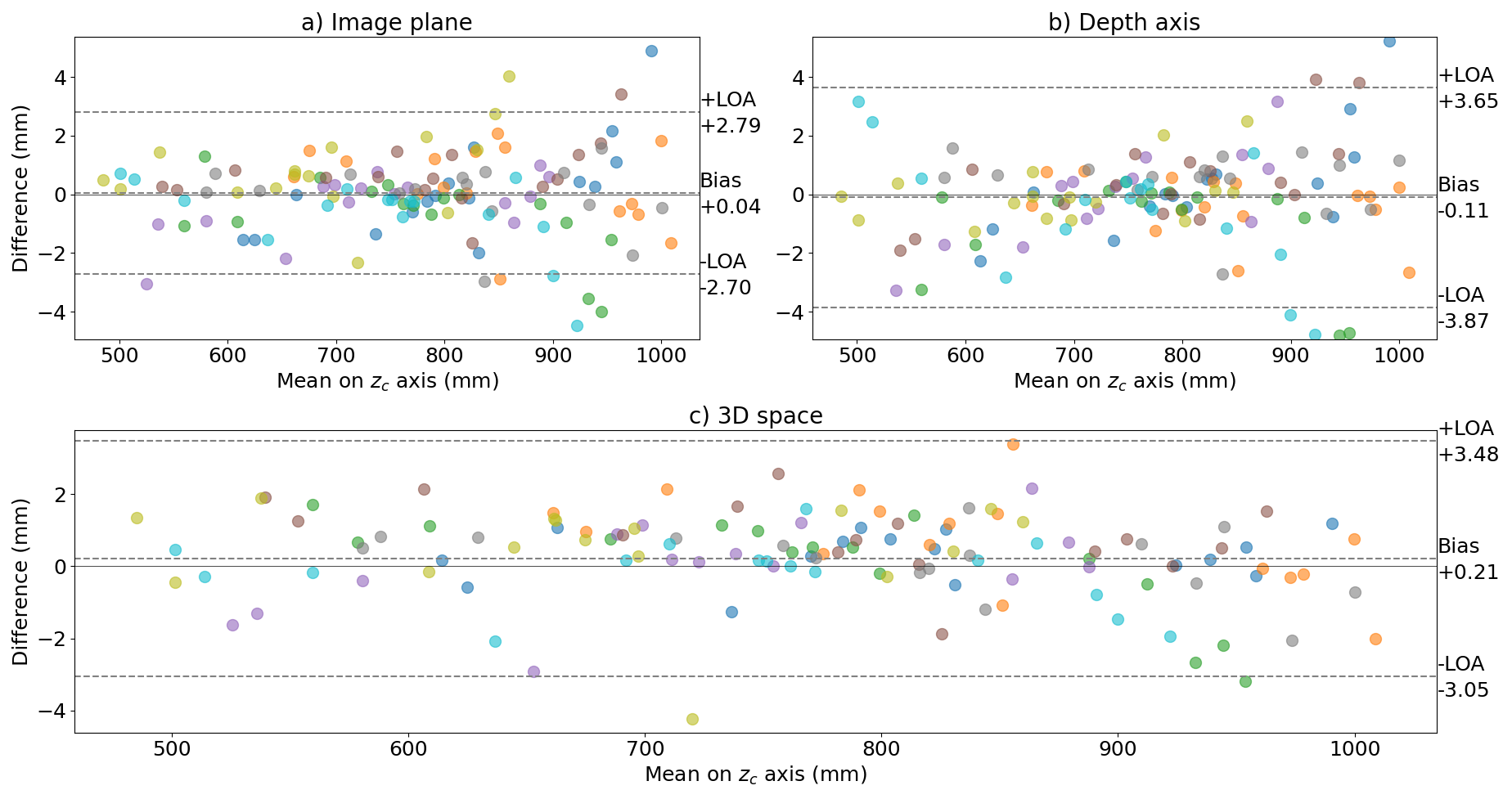}}
    \caption{Bland-Altman plot of the markers' positions expressed in the RGBD camera coordinate system, each marker being represented by a single point, distinguished by participant-specific colors. The differences were computed along the image plane (a), depth axis (b), and the mean of all axes (c).}
\label{fig:markers_bland_alt}
\end{figure}

The 3D trajectories of the markers showed consistent patterns across both motion capture systems (Figure \ref{fig:markers_3d}). This observation was further supported by the RMSE of the marker's trajectories across all trials and participants, averaging 3.32 ± 3.94\,mm. 
The Bland-Altman analysis (Figure\,\ref{fig:markers_bland_alt}) showed a uniform dispersion of data points across the plot, indicating a minimal influence of markers depth on the agreement between the two systems. Agreement within the image plane (Fig\,\ref{fig:markers_bland_alt}-a), formed by the $x_c$ and $y_c$ axes, outperformed the one along the depth axis, $z_c$, (Fig.\,\ref{fig:markers_bland_alt}-b) with narrower limits of agreement ((-2.70\,mm, 2.79\,mm) vs (-3.87\,mm, 3.65\,mm), respectively). In both cases, the bias remained close to zero.

\subsection{Biomechanical outcomes}
\begin{figure}\centerline{\includegraphics[width=\columnwidth]{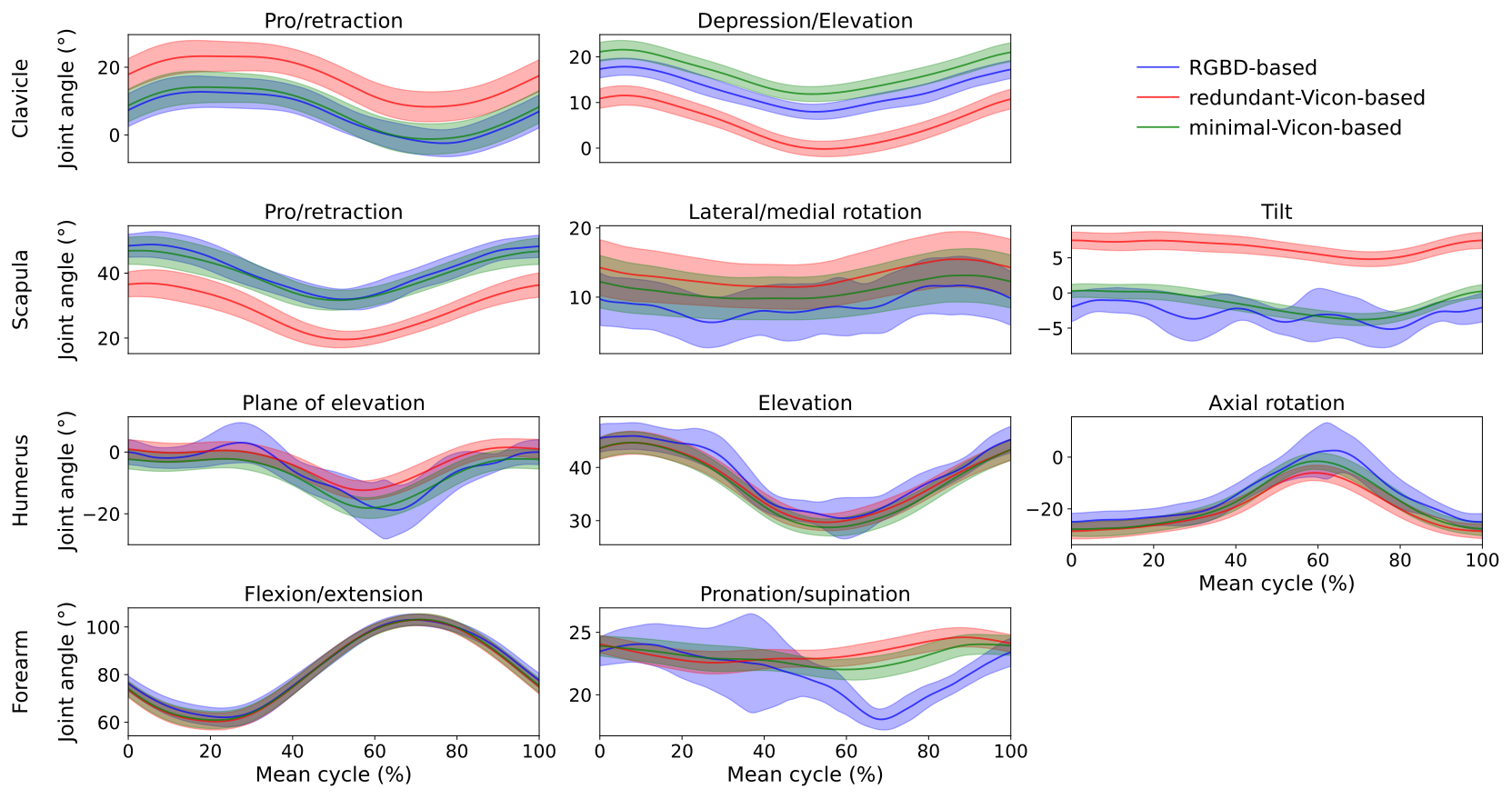}}
    \caption{Mean of 120 cycles of joint angles (solid line) with $\pm$ one standard deviation (shaded area) for a single participant for RGBD-based (blue), Vicon with redundancy (red), and minimal (green).}
\label{fig:q}
\end{figure}

\begin{figure}
\centerline{\includegraphics[width=\columnwidth]{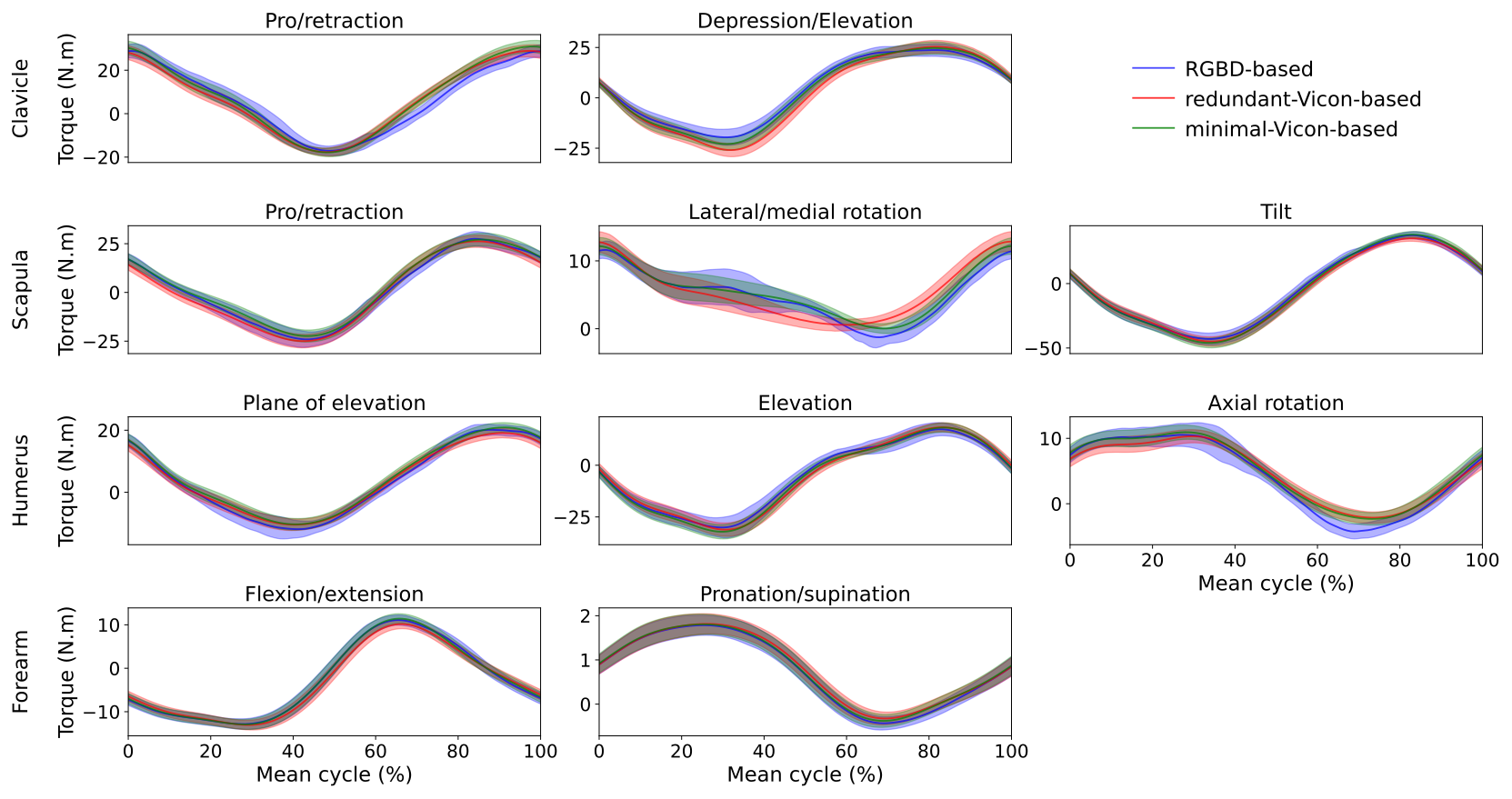}}
    \caption{Mean of 120 cycles of joint torques (solid line) with $\pm$ one standard deviation (shaded area) for a single participant for RGBD-based (blue), Vicon with redundancy (red), and minimal (green).}
\label{fig:tau}
\end{figure}

\begin{figure}
\centerline{\includegraphics[width=\columnwidth]{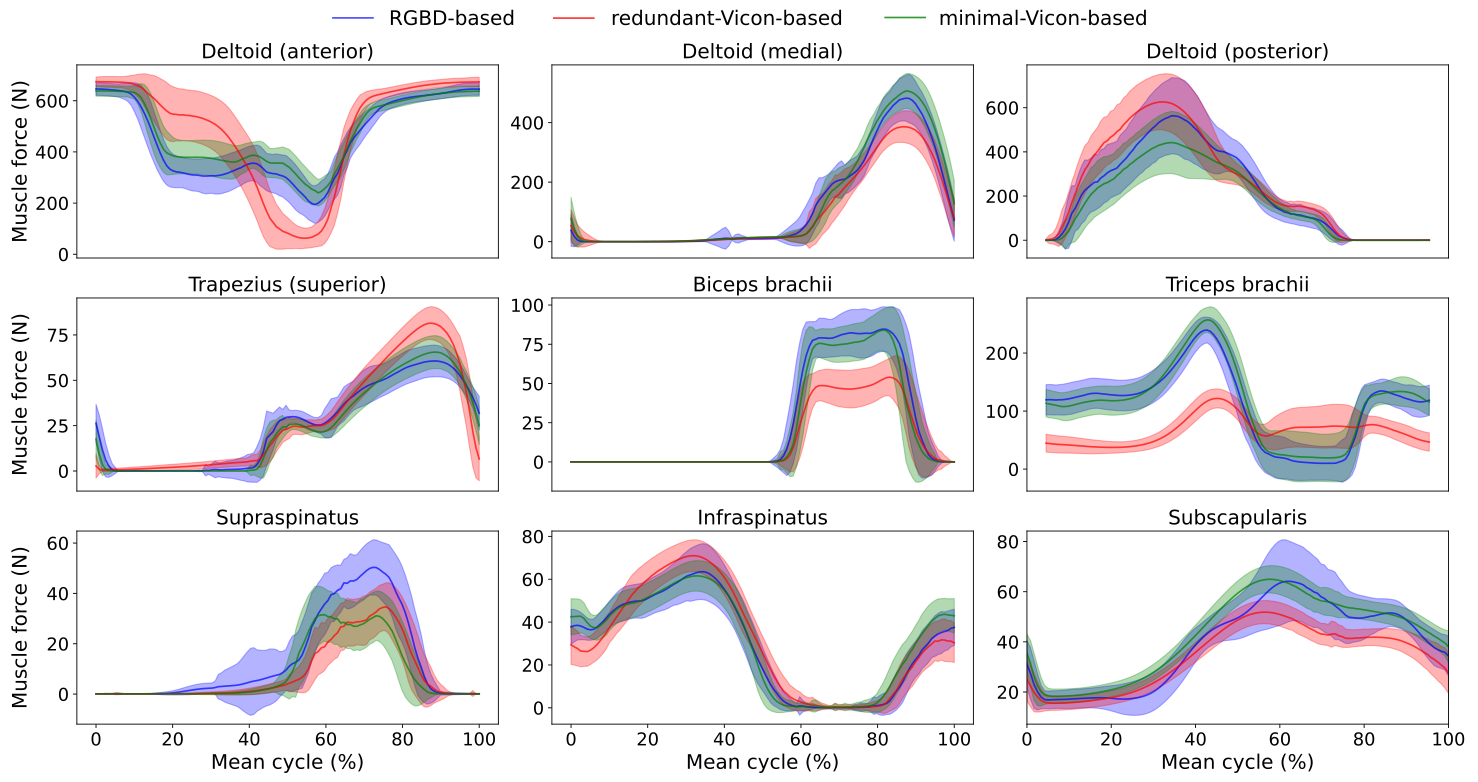}}
    \caption{Mean of 120 cycles of nine muscle forces (solid line) with $\pm$ one standard deviation (shaded area) for a single participant for RGBD-based (blue), Vicon with redundancy (red), and minimal (green).}
\label{fig:muscle}
\end{figure}

All three methods yielded personalized models with comparable body scaling factors across all participants (Table \ref{tab:scaling}).
The RMSD for joint angles for RGBDvMIN, showing an average difference of 2.34$\pm$2.79\degree, was approximately two times lower compared to RGBDvRED and MINvRED. A similar trend was observed for joint torques, with the RMSD for RGBDvMIN reaching 0.88$\pm$1.86\,N.m (Table \ref{tab:errors}). Additionally, the RMSD for muscle forces for RGBDvMIN, showing an average difference of 3.55$\pm$13.68\,N, was approximately three times lower compared to RGBDvRED or MINvRED (Table \ref{tab:errors}).
These results were supported by the Bland-Altman analysis, where the limits of agreement on joint angles (-4.53\degree~to 5.01\degree) and on joint torques (-1.70\,N.m, 1.40\,N.m) were approximately two times narrower for RGBDvMIN, compared to RGBDvRED and MINvRED. Likewise, the limits of agreement on the muscle forces for RGBDvMIN (-19.96\,N, 16.79\,N) were more than two times narrower than the two other comparisons. 
In all cases, the biases were close to zero for all comparisons, with the highest (but still low) value observed for muscle forces in the RGBDvRED comparison (-2.12\,N).
For each outcome, the analysis revealed a similar level of agreement when comparing the RGBD-based and minimal-Vicon methods to the redundant-Vicon method.
Figures\,\ref{fig:q}-\ref{fig:muscle} provide a comprehensive overview of joint angles, torques, and muscle forces during the mean cycle of a participant undergoing a resistance of 40\,W. Joint angles across all methods exhibited the same pattern, although some offsets are noticeable between redundant-Vicon and the other methods, especially on the sternoclavicular and acromioclavicular joints. Both methods with fewer markers showed almost no offset between each other. The RGBD-based method provided more noisy trajectories, especially for scapula rotation and tilt and forearm pronation/supination.
The joint torques provided by all three methods (Fig.\,\ref{fig:tau}) were nearly superimposed, although the effect of noisier joint angles was noticeable with a higher standard deviation area for the RGBD-based method.
The muscle forces showed almost the same onsets during the cycle, with visible offsets in amplitude for the redundant-Vicon method compared to the two other methods (Fig.\,\ref{fig:muscle}).

\begin{table}
    \caption{Scaling factors (mean along all axes for all participants) by segment for the three methods}
    \centering
    \begin{tabular}[c]{l c c c}
    \hline
    &  \multirow{2} * {RGBD} & \multicolumn{2}{c}{Vicon} \\
    Segment & & redundant & minimal \\ \hline
    Thorax & 1.06 $\pm$ 0.04 & 1.06 $\pm$ 0.05 & 1.07 $\pm$ 0.04  \\ 
Clavicle & 1.10 $\pm$ 0.09 & 1.08 $\pm$ 0.08 & 1.09 $\pm$ 0.08\\
Scapula & 1.21 $\pm$ 0.04 & 1.21 $\pm$ 0.05 & 1.21$\pm$ 0.05\\
Humerus & 0.98 $\pm$ 0.08 & 0.98 $\pm$ 0.07 & 0.97 $\pm$ 0.07\\
Forearm & 0.99 $\pm$ 0.06 & 1.01 $\pm$ 0.07 & 0.99 $\pm$ 0.05\\
\hline
    \end{tabular}
    \label{tab:scaling}
    \end{table}
    
\begin{table}
    \caption{Root Mean Square Deviation (RMSD) and one standard deviation (SD), along with Bland-Altman limit of agreement (LOA) and bias, of the joint angles and torques and muscle forces. Results are presented for both the RGBD-based and minimal-Vicon methods, compared with the redundant-Vicon method, and between the RGBD-based and minimal-Vicon methods}
    \centering
    \begin{tabular}{l l cc cc c}
    \hline
         & & RMSD & SD & \multicolumn{2}{c}{LOA} & Bias \\
         &  &  & & lower & upper & \\
         \hline
         \multirow{3}*{Joint angles (\degree~)} & RGBD vs redundant & 4.50& 2.85 & -12.13& 11.22& -0.46\\
& minimal vs redundant & 4.46& 0.84  & -13.28& 11.89& -0.70\\
& RGBD vs minimal & 2.34& 2.79  & -4.53& 5.01& 0.24\\
 \hdashline
\multirow{3}*{Joint torques (N.m)} & RGBD vs redundant & 2.26& 2.49 & -3.83& 5.07& 0.62\\
& minimal vs redundant & 2.25& 1.56  & -4.19& 5.73& 0.77\\
& RGBD vs minimal & 0.88& 1.86  & -1.70& 1.40& -0.15\\
 \hdashline
\multirow{3}*{Muscle forces (N)} & RGBD vs redundant & 10.70& 20.89 & -51.64& 47.40& -2.12\\
& minimal vs redundant & 10.60& 17.55  & -51.65& 50.58& -0.53\\
& RGBD vs minimal & 3.55& 13.68  & -19.96& 16.79& -1.58\\
\hline
                              
    \end{tabular}
    \label{tab:errors}
\end{table}

\begin{table}
\caption{Mean and standard deviation (SD) of each introduced delay per frame}
\centering
\begin{tabular}[c]{llcc}
\hline 
& & \multicolumn{2}{c}{Delay (ms)} \\
&  & mean & SD \\
\hline
\multirow{2}*{Tracking} & 2D tracking and correction & $7.64$ & $1.83$\\ 
& 3D correction & $2.71$ & $0.58$\\ 
\hdashline
\multirow{3}*{Biomechanical analysis} & Inverse kinematics & $0.47$ & $0.042$\\ 
  & Inverse dynamics & $0.31$ & $0.051$\\
& Static optimization and muscle force computation & $3.94$ & $1.54$\\ 
\hline
& \textbf{Total} &  $\boldsymbol{15.07}$ & $\boldsymbol{4.04}$\\
\hline 
\end{tabular}
\label{tab:delay}
\end{table}

\subsection{Time performance}
Table \ref{tab:delay} outlines the delays incurred by each task, ranging from RGBD tracking to biomechanical analysis. Due to the multiprocessing of the three areas, the total RGBD tracking delay across all participants was 10.35$\pm$2.41\,ms, corresponding to a tracking frequency of 94\,Hz, which exceeds the camera's FPS (60\,Hz). Once marker trajectories were retrieved, the delay introduced by the biomechanical analysis was approximately 4.72$\pm$1.63\,ms, corresponding to a solving rate of 211\,Hz, significantly surpassing the camera's FPS. The feedback delay was also affected by the markers' filtering using a live moving average method, which introduces a delay of nearly seven frames. Given that the system rate was fixed at 120 Hz, this resulted in a total delay of 58\,ms per frame. However, this delay does not impact the frequency of the algorithm, reaching a mean of around 63\,Hz (15.07$\pm$4.04\,ms Tab.\ref{tab:delay}).
Considering the delay introduced by the tracking, the biomechanical analysis and the marker filtering latency of the estimation from the recorded data was about 73.07$\pm$4.04\,ms.

\section{Discussion}
Our objective was to assess the viability of using a single RGBD camera for upper-limb biomechanical analysis.
We proposed a semi-automatic labeling algorithm and an efficient EMG-informed biomechanical pipeline to provide real-time feedback based on a single RGBD camera. 
Our results indicate that our approach yields accurate real-time 3D marker trajectories, enabling comprehensive upper-limb biomechanical assessments.
The provided user-friendly graphical interface simplifies the use of the tracking algorithm, while its single-camera setup makes it straightforward to install, making it suitable for clinical settings. Furthermore, our method allows recording from the front, enabling the capture of most upper limb bodies, including the scapula and the clavicle.
\subsection{Marker accuracy}
The 3D marker positions provided by RGBD-based mocap showed a good-to-strong level of agreement with those from Vicon-based mocap, as illustrated by the Bland-Altman analysis. 
The bias close to zero (-0.21 mm) and the limits of agreement within the range of (-3.05 mm, 3.48 mm) are lower than the errors induced by marker placement (up to 9 mm \cite{salvia2009precision}) or soft tissue artifact (up to 18 mm \cite{ancillao2021effect}).
Although the marker-to-camera distance did not much impact the level of agreement, wider limits of agreement were observed on the depth axis compared to the image plane.
This suggests that a higher part of the error may be introduced in the 3D space due to the depth image resolution or its misalignment with the RGB image. Such errors could potentially be mitigated by increasing the camera resolution or through one camera depth re-calibration \cite{darwish2017new}.

\subsection{Low propagated errors}
It is known that increasing the number of markers improves kinematics estimation accuracy, other than adding markers assessment can be made through numerical techniques, like the use of kinematic chains \cite{begon2008kinematics}. However, even with clinicians' expertise in marker placement, excessive marker placement can be too time-consuming. Our marker count with the RGBD camera aligns with other clinical upper limb marker sets, using, for instance, 7 \cite{vanezis2015reliability}, 8 \cite{blaszczyszyn2023quantitative} or 12 \cite{noble2018practical} markers. Our evaluation revealed that, for each biomechanical outcome, the level of agreement, as indicated by the bias and limits of agreement, was similar when the same marker set was used. When comparing both minimal methods, the agreement level was twice as strong as that observed between the minimal and redundant methods, emphasizing the greater influence of the marker set on biomechanical outcomes compared to using two distinct motion capture methods.
Implementing a marker cluster on the thorax, as done in Noble \& al. \cite{noble2018practical}, has the potential to mitigate incorrect initial thorax orientation, thereby improving joint angle accuracy on attached segments.
That being said, the comparison between the RGBD and minimal-Vicon methods showed a strong agreement, with limits of agreement on joint angles ranging from -4.53\degree~to 5.01\degree. The error in joint angle estimation between RGBD and redundant-Vicon methods ($\sim$5\degree) was lower than those found in the literature for the upper limb using IMU systems ($\sim$30\degree~\cite{henschke2022assessing}) or markerless systems ($\sim$23\degree~\cite{lahkar2022accuracy}).
These values fall within the variability typically observed when using kinematics reconstruction methods, which can yield errors of up to 10\degree~between sessions in upper-limb analysis for children with cerebral palsy \cite{jaspers2011reliability}.
Similarly, Uchida \& al. \cite{uchida2022conclusion} reported errors in joint torque estimation of up to 26.6\,N.m (17$\%$ of the peak) whereas in our study, the limits of agreement of joint torque varied from -1.70\,N.m to 1.40\,N.m (about 7$\%$ of the highest values of joint torques, i.e. $\pm$25\,N.m.).
We observed higher agreement in the computation of joint torques compared to the estimation of muscle forces. This can be attributed to differences in joint kinematics, which impact muscle force estimation through variations in moment arms or muscle length. This is demonstrated by Blache and Begon \cite{blache2017influence}. However, despite these differences, all methods produced similar muscle force patterns, with the resulting onset potentially being the same depending on the threshold used.

\subsection{Latency}
Waltermate \& al. \cite{waltemate2016impact} showed that motor performance and perception are affected by latencies above 75\,ms . The ultimate goal of this work is to facilitate real-time biofeedback to guide rehabilitation motion, making the delay introduced by our system a critical factor. We developed an efficient marker labeling pipeline based on Python multiprocessing, enabling the retrieval of labeled markers with a delay of 10.6\,ms (94\,Hz), significantly exceeding the camera's frame rate. This ensures that our algorithm can provide labeled markers promptly, even using cameras with a higher frame rate.
For the biomechanical pipeline, we used C++-based libraries for musculoskeletal modeling (biorbd) and static optimization (acados) providing outcomes with a delay of 4.7\,ms (213\,Hz), significantly exceeding the camera's frame rate. However, the need to filter the marker trajectories induces a significant delay of 58\,ms per frame. As a result, the feedback can be delivered with a delay of 73.07\,ms, which is of the same order as the 75\,ms target. This frequency could potentially be further improved by executing marker retrieval and biomechanical analysis in parallel. Additionally, marker filtering could be performed using a Kalman filter, which should drastically reduce the delay to nearly 0\,ms \cite{spincemaille2008kalman}.

\subsection{Clinical implication}
For our novel low-cost motion capture system, we have designed a graphical interface to simplify its utilization in various settings, from research to clinical applications. 
The tracking algorithm is semi-automatic, facilitating setup procedures; nonetheless, manual labeling remains necessary for the initial frame. 
To enhance efficiency, iterative closest point methods could be integrated to automatically label the initial frame using a pre-implemented model \cite{ying2009scale}. Moreover, we can introduce a calibration step to establish a z-vertically oriented coordinate system directly from the camera. This can be achieved either by employing a 3D shape with white markers (like the calibration 5-marker wand used by the Vicon system) or by using the inertial sensors integrated into the camera.
The next step in this work is to evaluate the system in a clinical context with the involvement of clinicians.\\

This study has some limitations. Foremost, the camera operates at a non-conservative 60 Hz frame rate, which could potentially impact time-dependent analyses. However, this limitation can be effectively managed as the camera provides the image count, allowing the detection and handling of any discontinuities using interpolation methods.
Furthermore, synchronizing the camera with other devices such as EMG or force plates can be challenging due to the non-research-oriented nature of the chosen camera. External tools like biosiglive \cite{ceglia2023biosiglive} can assist in retrieving data from various sources, though further development is required. A future objective for our low-cost motion capture system could involve developing an embedded system to facilitate data dissemination more reliably. Future work will leverage our automatic labeling algorithm to provide annotated depth images, which can be used to train a new machine learning model for markerless pose estimation based on bony landmarks.

\section{Conclusion}
We showed that using a single low-cost RGBD camera and white skin markers enables upper limb biomechanical analysis comparable to an optoelectronic system. By developing a semi-automatic labeling algorithm and an EMG-informed biomechanical pipeline, we provided real-time feedback based on a single RGBD camera. Our approach provides accurate real-time 3D marker trajectories, facilitating comprehensive upper-limb biomechanical assessments. With its user-friendly interface and single-camera setup, our system is well-suited for clinical settings.
We observed strong agreement between the RGBD and Vicon methods when considering the same amount of recorded markers. However, we highlighted the impact of the lack of marker redundancy on inverse kinematics, leading to notable differences using both RGBD and minimal-Vicon systems compared to the redundant Vicon system.
Overall, our findings underscore the potential of RGBD cameras as a cost-effective alternative for upper-limb biomechanical analysis, with opportunities for further refinement and enhancement in future research.

\section{Conflict of interest statement}
The authors declare that they have no known competing financial interests or personal relationships that could have appeared to influence the work reported in this paper.

\section{Authors contributions}
AC, MB and LS: conceptualization. AC and KF: data curation. AC, MB and LS: methodology. AC and KF: software. MB and LS: supervision. AC: writing-original draft. MB and LS: writing-review \& editing. AC and MB: Funding acquisition. All authors have read and agreed to the published version of the manuscript.

\section{Funding}
This work was supported by the Natural Sciences and Engineering Research Council of Canada (NSERC) through the CREATE OPSIDIAN program, the discovery grant RGPIN/04978-2019, and the Fonds de recherche du Québec – Nature et technologies (FRQNT).

\bibliographystyle{alpha}
\bibliography{sample}

\end{document}